\documentclass[11pt]{article}

\usepackage[margin=1in]{geometry}
\usepackage{amsmath, amssymb}
\usepackage{booktabs}
\usepackage{graphicx}
\usepackage{subcaption}
\usepackage{hyperref}
\usepackage{siunitx}
\usepackage{enumitem}
\usepackage{tikz}
\usetikzlibrary{arrows.meta,positioning,calc,fit,backgrounds,decorations.pathmorphing}

\hypersetup{
  colorlinks=true,
  linkcolor=blue,
  urlcolor=blue,
  citecolor=blue
}

\title{Imputing Missing Near-Surface Air Temperature using Graph Attention Networks with Distance-Aware Edge Features}
\author{Emma C. Weber, Dominique Guillot, Daniyar Abykhanov, Julien Emile-Geay}
\date{\today}

\begin{document}
\maketitle

\begin{abstract}
    We introduce a graph attention network for reconstructing near-surface air temperature fields from incomplete observations. The model represents each grid cell as a node connected to its $k$ nearest neighbors, with attention weights that depend on node features, distance, and land–ocean transitions. This graph representation avoids the polar distortions associated with treating the globe as a flat latitude–longitude image, as most deep learning reconstruction methods do. This is particularly relevant for historical temperature records, which extend back to the mid-nineteenth century but remain incomplete, with the largest gaps early in the record and at high latitudes. We train our model on CMIP6 CESM2 historical simulations masked with the time-varying HadCRUT5 observational coverage from 1850 to 2014. We then compare our approach with ordinary kriging, GraphEM, and the LaMa inpainting network on a held-out realization. The graph model achieves the lowest spatial RMSE (0.997 K), 10.1\% lower than LaMa and approximately 31\% lower than kriging and GraphEM. The largest gains occur in the Arctic and Antarctic, while LaMa remains slightly more accurate in the densely observed tropics. The graph model’s error also falls faster as coverage increases, so its advantage over LaMa grows by more than a third from the 1850s to the 2000s. An ablation study shows that the model’s skill comes mainly from the geometric graph construction rather than from auxiliary physical covariates.
\end{abstract}

\section{Introduction}
\label{sec:intro}

Observational records of climate fields are sparse, and the sparsity increases going back in time \cite{Morice2021}. Surface air temperature is among the best-observed fields and reaches back to the mid-nineteenth century. Fewer than half of all grid cells are observed on average, the gaps are spatially clustered and persistent, and some locations are never observed at all.

How these gaps are filled affects conclusions drawn from the data. The Arctic has warmed nearly four times faster than the global mean since 1979 \cite{Rantanen2022}, and it is also among the least observed regions in the instrumental record. Coverage bias in gridded products has been shown to alter estimates of recent warming trends \cite{cowtanway2014}. Reconstruction methods are therefore evaluated not only on average error but on their behavior in the regions where observations are scarcest.

Gaps in the instrumental record have traditionally been filled with statistical methods. One family interpolates from nearby observations. GISTEMP weights station anomalies by their distance from each grid point \cite{Hansen2010, Lenssen2019}, while kriging and Gaussian-process regression estimate missing values from a fitted spatial covariance function \cite{cressie1993statistics}. These underlie Berkeley Earth \cite{Rohde2013}, the Cowtan--Way infilling of HadCRUT \cite{cowtanway2014}, and the HadCRUT5 statistical analysis \cite{Morice2021}. A second family projects the field onto a reduced basis of empirical orthogonal functions (EOFs), estimated either from a well-observed reference period \cite{Kaplan1998, Smith2008}, or iteratively from the incomplete data  \cite{BeckersRixen2003}. A third family estimates the covariance structure and missing values jointly by expectation-maximization \cite{Schneider2001, Guillot2015} and have been applied directly to HadCRUT \cite{Vaccaro2021}. A fourth family, Bayesian hierarchical models such as BARCAST \cite{TingleyHuybers2010}, models the field jointly in space and time and yields a full posterior distribution over the reconstruction. These methods are interpretable and well understood, but many rely on linear, largely Gaussian structure, tend to smooth extremes in data-sparse regions, and most must be refit for each new dataset or coverage pattern.

Machine learning approaches learn the structure of climate fields from complete climate model output and apply it to sparse observations. Neural networks have been incorporated into operational products, such as improving land interpolation in NOAAGlobalTemp \cite{Huang2022}. Most work, however, treats reconstruction as image inpainting on the latitude--longitude grid. Kadow et al.\ \cite{Kadow2020} applied partial convolutions to HadCRUT4, an approach extended to climate extremes by Plésiat et al.\ \cite{Plesiat2024}. Bochow et al.\ \cite{Bochow2025} applied LaMa \cite{Suvorov2022}, a Fourier-convolution inpainting network, and reported a 21\% lower spatially averaged site-wise RMSE than kriging. Convolutional autoencoders are also used to fill gaps in satellite sea surface temperature records \cite{Barth2020}. All of these operate on a regular pixel lattice. Bochow et al.\ note that convolutional networks do not capture the spherical geometry of the Earth, producing artifacts at the image edges and degrading performance in the polar regions, and suggest graph neural networks or spherical convolutions as possible remedies \cite{Bochow2025}. They outlined two directions for improvement. The first, generative and spatiotemporal modeling, has since been pursued with diffusion models that condition on sparse observations to produce probabilistic, temporally consistent reconstructions \cite{Qian2026}. The second is to adopt architectures that respect the spherical geometry of the Earth. Graph neural networks have been developed for spatial interpolation more generally \cite{Appleby2020}, but to our knowledge they have not been applied to the reconstruction of global historical temperature fields.

We pursue the second direction and introduce a graph neural network for this task. A graph neural network operates on data defined on a graph, which is a set of nodes joined by edges \cite{Scarselli2009, Kipf2017}. In each layer, every node updates its state by combining information from the nodes it is connected to; stacking several layers lets information propagate across the graph. This resembles kriging, in which an estimate is a weighted combination of nearby values, except that the weights are learned from data and the combination is nonlinear. In our setting, grid cells become nodes, and each node is connected to its $k$ nearest neighbors. This allows the Earth's geometry to be represented directly, since locality is defined by physical distance rather than by grid index, connectivity wraps the dateline, and the poles are connected to their surroundings rather than treated as distorted edges of an image. Spherical convolutions offer another route to the same goal but remain considerably more expensive than standard convolutions \cite{Bochow2025}, whereas a graph neural network operates directly on the native grid at a cost that scales with the number of edges. The graph also provides a natural interface for auxiliary information. Physical covariates such as elevation and land--sea structure attach to nodes, and geometric quantities such as distance attach to edges. This allows new inputs to be added without changing the architecture. We use a distance-aware GATv2 \cite{brody2022gatv2}, in which each node assigns a weight to each of its neighbors based on their features and the distance between them. The model can then learn to weight a distant ocean neighbor above a closer neighbor across a coastline through training. The resulting model reconstructs a full historical realization in seconds.

To test whether a graph-based representation improves reconstruction under realistic conditions, we use CMIP6 CESM2 historical simulations \cite{eyring_cmip6_2016} as ground truth and mask them with the time-indexed HadCRUT5 record. Every month carries exactly the coverage available in that historical calendar month, including the persistent polar gaps and the sparse coverage of the nineteenth century. Eighteen ensemble realizations are used for training, one is held out for validation and model selection, and one is held out for evaluation. Rather than comparing against every published product, we select one representative from each of the main families of reconstruction methods. Kriging represents interpolation from nearby observations since it is the statistical core of Berkeley Earth, the Cowtan--Way infilling, and the HadCRUT5 statistical analysis. GraphEM represents expectation-maximization methods with a spatial model, and has been applied directly to HadCRUT \cite{Vaccaro2021}. Because it also operates on an explicit graph, we supply it with the same graph as our model, so that the comparison isolates a learned nonlinear model from a linear Gaussian one given identical spatial structure. LaMa represents one of the latest deep-learning inpainting approaches to global temperature reconstruction \cite{Bochow2025}. Finally, all four methods are evaluated under a single protocol: the same test realization, the same time-varying masks, the same seasonal normalization computed from the training simulations, and the same latitude-weighted metrics evaluated on a common set of unobserved grid cells.

We show that the graph attention model attains the lowest overall error of the four methods, with the largest margins at high latitudes, where observational coverage is poorest and where lattice-based methods are weakest. However, in the tropics where coverage is dense and the grid is least distorted, LaMa and GraphEM are slightly more accurate. At the 330 grid cells never observed in HadCRUT5, the graph attention model's error is 13\% lower than LaMa's and 27\% lower than kriging's. An ablation study shows that this advantage comes primarily from the geometric construction of the graph. We further show that the graph model and LaMa respond differently to observational density. Both improve as coverage increases over the record, but the error of the graph attention model declines faster. This propagates as an error difference between the two models that widens by more than a third between the 1850s and the 2000s. The graph attention model therefore makes better use of each additional observation, which matters as historical archives continue to be digitized and new observations are added to the record.

\section{Methods}
\label{sec:methods}

\subsection{Problem Formulation}

Let $X \in \mathbb{R}^{T \times N}$ be a temperature field over $T$ months and $N$ grid cells, and let $M \in \{0,1\}^{T \times N}$ be the observation mask, where $M_{t,i} = 1$ if cell $i$ is observed at month $t$. All fields are standardized as described in Section~\ref{sec:data}, so $X$ is expressed in units of the local seasonal standard deviation.

The model receives the masked field $X \odot M$ alongside $M$ and predicts $\hat{X}$, where $\odot$ denotes entry-wise (Hadamard) multiplication, $(X \odot M)_{t,i} = X_{t,i} M_{t,i}$. Unobserved entries are therefore set to zero, which in the standardized representation corresponds to the local climatological mean for that calendar month. Training minimizes the mean squared error evaluated strictly at unobserved cells:
\begin{equation}
  \mathcal{L} = \frac{\sum_{t=1}^{T}\sum_{i=1}^{N} (1 - M_{t,i})(\hat{X}_{t,i} - X_{t,i})^2}
                     {\sum_{t=1}^{T}\sum_{i=1}^{N} (1 - M_{t,i})} .
\end{equation}
Observed cells contribute no gradient, ensuring the model is not rewarded for reproducing provided values.

\subsection{Graph Neural Networks}
\label{sec:gnn_background}
 
A graph consists of a set of \emph{nodes} and a set of \emph{edges}, where each edge connects a pair of nodes. In our setting each node is a grid cell, and an edge from node $j$ to node $i$ indicates that information at $j$ may be used to estimate the value at $i$. We write $\mathcal{N}(i)$ for the \emph{neighborhood} of node $i$, which is the set of nodes with an edge connecting to $i$.
 
Both nodes and edges carry \emph{features}, which are fixed-length vectors of numbers describing them. A node feature vector $\mathbf{x}_i$ describes a location, for example its temperature, whether it is observed, and its elevation. An edge feature vector $\mathbf{e}_{ij}$ describes the relationship between two locations, for example the distance between them.
 
A graph neural network (GNN) learns a function on such graphs through repeated rounds of \emph{message passing}. Each node holds a hidden representation $\mathbf{h}_i$, a vector initialized from its features. Each layer performs three steps. First, every node receives a message from each of its neighbors, computed from the neighbor's representation and the connecting edge's features. Second, the node aggregates these messages into a single vector, typically by a weighted sum. Third, the node updates its own representation using the result. The same learned functions are applied at every node and every edge, so the number of parameters does not depend on the size of the graph. After one layer a node's representation reflects its immediate neighbors, and after $L$ layers it reflects all nodes within $L$ edges. Stacking layers thus allows information from observed regions to propagate progressively into data gaps. Conceptually, this resembles distance-weighted interpolation, except that the weights, and the way information is combined, are learned from data rather than prescribed.
 
\begin{figure}[htbp]
\centering
  \includegraphics[width=\linewidth]{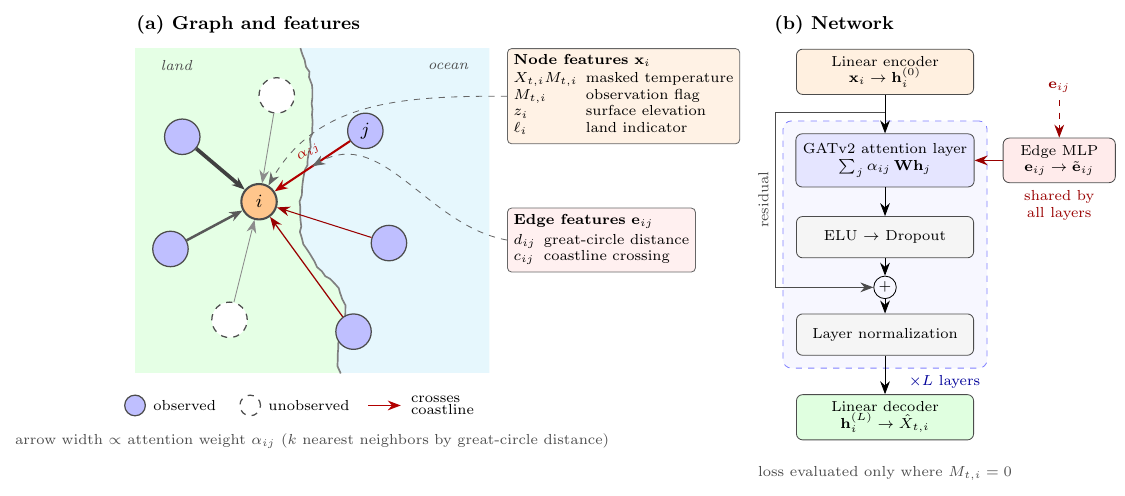}
\caption{Schematic of the graph attention network. (a) Each grid cell is a node connected to its $k$ nearest neighbors by great-circle distance. Node $i$ (orange) is unobserved at month $t$, and its neighbors may be observed (blue) or unobserved (dashed). Every node carries four features (masked temperature, observation flag, elevation, land indicator) and every edge carries two (distance and whether the edge crosses a coastline). The network learns an attention weight $\alpha_{ij}$ for each edge, indicated by arrow width, which determines how strongly neighbor $j$ contributes to the estimate at $i$. (b) Node features are encoded into a hidden representation, refined by $L$ GATv2 layers that each use the same learned edge embedding, and decoded to a temperature estimate at every node.}
\label{fig:gat_schematic}
\end{figure}

The simplest GNNs weight all neighbors equally or by a fixed rule. Graph attention networks (GATs) \cite{velivckovic2018gat} instead learn the weights. For each edge $j \to i$, the network computes a score that measures how relevant neighbor $j$ is to node $i$, and normalizes the scores over the neighborhood so that they are positive and sum to one. These normalized scores, the \emph{attention coefficients} $\alpha_{ij}$, are then used as the weights when summing the neighbors' messages. Because the coefficients are recomputed from the current representations, a node may rely on different neighbors in different months. For example, it may draw on observed upstream cells when they are available and on more distant cells when they are not. We use the GATv2 variant \cite{brody2022gatv2}, described in Section~\ref{sec:architecture}, which makes this neighbor ranking more flexible. Figure~\ref{fig:gat_schematic} summarizes the full pipeline for our configuration.

\subsection{Graph Construction}
\label{sec:graph}
 
We represent the grid as a graph with one node per grid cell. The graph lets physical information enter the model in two places: as attributes of locations (node features) and as attributes of the connections between them (edge features). Attributes of both kinds are standard in graph neural networks for physical systems \cite{Lam2023}, but since graph neural networks have not previously been applied to the reconstruction of historical temperature fields, the features below are a first exploration of which physical information is useful for this task rather than an optimized set. Their individual contributions are assessed in Table~\ref{tab:ablation}.
 
\paragraph{Node features.} Each node carries four features. The two time-varying features are the masked standardized temperature $X_{t,i} M_{t,i}$ and the observation indicator $M_{t,i}$. The two static features are surface elevation and a binary land indicator. Supplying $M$ explicitly is necessary because zero represents a valid anomaly value. Without the mask channel, the model cannot distinguish an observed cell at its climatological mean from an unobserved cell.
 
\paragraph{Edges.} Edges are determined by physical geometry rather than by adjacency in the latitude--longitude array. For each node $i$, we compute the great-circle distance to every other node using the haversine formula with Earth radius $R = 6371$\,km
\begin{equation}
  d_{ij} = 2R \arcsin \sqrt{ \sin^2\!\Big(\tfrac{\Delta\phi}{2}\Big)
           + \cos\phi_i \cos\phi_j \sin^2\!\Big(\tfrac{\Delta\lambda}{2}\Big) } ,
\end{equation}
where $\phi$ denotes latitude, $\lambda$ denotes longitude, and we connect $i$ to its $k$ nearest neighbors. Because the nearest-neighbor relation is not symmetric, as node $j$ may be among the $k$ nearest neighbors of $i$ without $i$ being among those of $j$, we symmetrize the edge set by adding the reverse of every edge and removing duplicates, so that most nodes have slightly more than $k$ neighbors.

The pole rows of the grid require additional treatment. Each consists of 72 grid points at the same physical location, at zero great-circle distance from one another. Under plain nearest-neighbor selection, each pole node would choose copies of itself as its neighbors, and each pole row would form a component disconnected from the rest of the graph. We therefore skip coincident points when choosing neighbors, so that every node is connected to its $k$ nearest distinct locations. All cells of the adjacent $85^\circ$ row are equidistant from the pole, so ties are broken by longitude, and the pole node at longitude $\lambda$ connects to the $k$ cells of the adjacent row nearest to $\lambda$. Away from the poles there are no coincident points, and the graph is exactly the symmetrized $k$-nearest-neighbor graph. For $k = 16$ the graph has $|E| = 46{,}800$ directed edges. Every node has at least $k$ neighbors, with a mean of $17.6$ and a maximum of $36$. The graph is static, constructed once and shared across all timesteps.
 
\paragraph{Edge features.} Each edge carries distance $d_{ij}$ (standardized to zero mean and unit variance across the edge set) and a binary land--ocean transition indicator equal to one when $i$ and $j$ lie on opposite sides of a coastline. The distance lets the attention weights depend on how far apart two locations are, rather than treating all $k$ neighbors alike. The transition indicator supplies an explicit signal at boundaries where temperature covariance changes due to the contrasting thermal inertia of land and ocean.

\subsection{Network Architecture}
\label{sec:architecture}
 
The network consists of an encoder, a stack of attention layers, and a decoder (Figure~\ref{fig:gat_schematic}b). Specific layer widths and settings are listed in Appendix~\ref{app:hyperparameters}.
 
\paragraph{Encoder.} A linear layer maps each node's four input features to a higher-dimensional hidden representation $\mathbf{h}_i$, giving the network room to form combinations of the inputs. Separately, a small multilayer perceptron (MLP) maps the two edge features to an edge embedding $\tilde{\mathbf{e}}_{ij}$. The same edge embedding is reused in every attention layer and at every timestep, since the geometry it encodes does not change.
 
\paragraph{Attention layers.} The body of the network is a stack of GATv2 \cite{brody2022gatv2} layers, each conditioned on the edge embedding. For a node $i$ with neighborhood $\mathcal{N}(i)$, the attention coefficients are
\begin{equation}
  \alpha_{ij} = \frac{\exp\!\big(\mathbf{a}^{\top}\,\operatorname{LeakyReLU}
    (\mathbf{W}_{\!l}\mathbf{h}_i + \mathbf{W}_{\!r}\mathbf{h}_j
     + \mathbf{W}_{\!e}\tilde{\mathbf{e}}_{ij})\big)}
  {\sum_{q \in \mathcal{N}(i)} \exp\!\big(\mathbf{a}^{\top}\,\operatorname{LeakyReLU}
    (\mathbf{W}_{\!l}\mathbf{h}_i + \mathbf{W}_{\!r}\mathbf{h}_q
     + \mathbf{W}_{\!e}\tilde{\mathbf{e}}_{iq})\big)} ,
\end{equation}
where $\mathbf{W}_{\!l}$, $\mathbf{W}_{\!r}$, and $\mathbf{W}_{\!e}$ are learned matrices that project the receiving node, the sending node, and the edge into a common space, and $\mathbf{a}$ is a learned vector that converts the result into a scalar score. The softmax function makes the coefficients positive and sum to one over the neighborhood, so the update
\begin{equation}
  \mathbf{h}_i' = \sum_{j \in \mathcal{N}(i)} \alpha_{ij}\,\mathbf{W}_{\!r}\mathbf{h}_j
\end{equation}
is a learned weighted average of the neighbors' representations. Each layer uses multiple attention heads, which are several independent sets of attention coefficients computed in parallel, allowing the layer to attend to different aspects of the neighborhood simultaneously. The outputs of the heads are averaged, which keeps the hidden width constant from layer to layer.
 

Each attention layer is wrapped as
\begin{equation}
  \mathbf{h}^{(\ell+1)} = \operatorname{LayerNorm}\Big(\mathbf{h}^{(\ell)}
    + \operatorname{Dropout}\big(\operatorname{ELU}(\operatorname{GATv2}(\mathbf{h}^{(\ell)}))\big)\Big),
\end{equation}
which combines four standard components. Dropout randomly sets a fraction of the hidden values to zero during training, with the fraction set by a dropout probability. This prevents the network from relying on any single pathway and reduces overfitting to the training realizations. The residual connection adds the layer's input $\mathbf{h}^{(\ell)}$ to its output, so each layer learns a correction to the current representation rather than a replacement, which stabilizes training of deeper networks. Finally, layer normalization rescales each node's hidden vector to zero mean and unit variance, keeping values in a consistent range across layers.
 
\paragraph{Decoder.} A final linear layer maps each node's representation to a single standardized temperature estimate $\hat{X}_{t,i}$, which is converted to kelvin for evaluation. 
 
\subsection{Baselines}

The methods reviewed in Section~\ref{sec:intro} include interpolation from nearby observations, reduced-basis (EOF) methods, joint covariance estimation by expectation-maximization, Bayesian hierarchical models, and machine-learning inpainting on the latitude--longitude grid. Rather than comparing against every individual product, we select one representative from each of the three families most directly relevant to our approach, and evaluate all of them under identical conditions, using the same seasonal standardization computed from the training realizations, the same observation masks, and the same evaluation metrics. Hyperparameter choices are given in Appendix~\ref{app:baselines}.
 
\paragraph{Ordinary kriging.} Kriging \cite{cressie1993statistics} estimates each missing value as a weighted linear combination of observed values. The weights are derived from a variogram, a fitted function describing how the expected squared difference between two values grows with the distance between them, and are chosen to give the minimum-variance unbiased linear estimate under that model. We implement kriging via PyKrige \cite{murphy2022pykrige} with a spherical variogram fitted independently at each timestep. Kriging represents the interpolation family and is the statistical core of several widely used products, including Berkeley Earth \cite{Rohde2013}, the Cowtan--Way infilling of HadCRUT \cite{cowtanway2014}, and the HadCRUT5 statistical analysis \cite{Morice2021}. It is also the reference against which Bochow et al.\ \cite{Bochow2025} evaluated LaMa, allowing our results to be placed alongside theirs. Simpler distance-weighting schemes such as that of GISTEMP \cite{Hansen2010, Lenssen2019} belong to the same family but use prescribed weights, whereas kriging fits its weights to the data and accounts for redundancy among clustered observations. We therefore treat kriging as the stronger representative of this family. Like our model, kriging uses only observations from the same month, so differences between the two reflect how the weights are obtained.
 
\paragraph{GraphEM.} GraphEM \cite{Guillot2015} models the climate field as a multivariate Gaussian whose precision (inverse covariance) matrix is sparse, with nonzero entries permitted only between nodes connected in a prescribed graph. It alternates between estimating the covariance structure from the currently filled-in field and re-imputing the missing values from it, an expectation-maximization procedure that also underlies RegEM \cite{Schneider2001}. GraphEM represents the expectation-maximization family. Compared to similar approaches, GraphEM constructs an explicit spatial model for the field and is closer to our GNN approach as both methods operate on a graph. In pseudoproxy experiments, it was shown to reduce reconstruction error relative to RegEM with truncated total least squares regularization at nearly every location, by about 43\% on average, and tied the improvement to its sparse, local representation of the covariance structure. GraphEM has also been applied directly to HadCRUT \cite{Vaccaro2021}. Guillot et al.\ further showed that simple distance-based neighborhood graphs capture the spatial dependence of temperature fields well. We therefore supply GraphEM with the identical $K$-nearest-neighbor haversine graph used in our model, so that differences in skill reflect how each method uses the graph (a linear Gaussian model versus a learned nonlinear one) rather than which graph it is given. Unlike our model, GraphEM is transductive, meaning that it estimates its covariance directly from the target field being reconstructed, rather than learning it in advance from training data, and can draw on information from all timesteps of that field.
 
\paragraph{LaMa.} LaMa \cite{Suvorov2022} is an image inpainting network that uses Fourier convolutions, operations that mix information across the entire image in a single layer, together with adversarial, perceptual, and reconstruction losses. We adopt the implementation of Bochow et al.\ \cite{Bochow2025} configured for climate field reconstruction (\texttt{lama-fourier}). LaMa represents the machine-learning inpainting family and is, to our knowledge, the strongest published deep-learning method for global temperature reconstruction, having been reported to outperform both kriging and the partial-convolution approach of Kadow et al.\ \cite{Kadow2020}. Because LaMa treats the field as a flat image on the latitude--longitude grid, it is the natural comparison for testing whether representing the Earth's spherical geometry explicitly, as our graph does, improves reconstruction, particularly at high latitudes and across the dateline. Fields are encoded as 8-bit greyscale images, imposing an absolute quantization floor of $0.075$\,K on achievable error (derivation in Appendix~\ref{app:quantization}). LaMa is trained for 100 epochs ($356{,}400$ optimizer steps, $6.4\times$ the $55{,}700$ optimizer steps of our graph model).
 
\subsection{Evaluation Metrics}
 
We report spatial RMSE and mean site-wise RMSE \cite{Kadow2020, Bochow2025}, which differ in their order of temporal and spatial aggregation. Both metrics are computed in kelvin after denormalization, excluding known observed cells at each timestep.
 
Let $m_{t,i}$ be the observation mask, $\Omega_t = \{i \mid m_{t,i} = 0\}$ the unobserved cells at month $t$, and $w_i = \cos(\phi_i)$ the area weight at latitude $\phi_i$. The spatial RMSE at timestep $t$ is defined as
\begin{equation}
\operatorname{RMSE}_{\mathrm{spatial}}(t) =
\left(
\frac{\sum_{i\in\Omega_t} w_i (\hat{x}_{t,i}-x_{t,i})^2}
     {\sum_{i\in\Omega_t} w_i}
\right)^{1/2},
\end{equation}
and its record-wide mean is denoted $\operatorname{SpatialRMSE} = T^{-1}\sum_{t=1}^{T}\operatorname{RMSE}_{\mathrm{spatial}}(t)$.
 
For each grid cell $i$, let $\mathcal{T}_i=\{t \mid m_{t,i}=0\}$ be the set of timesteps where cell $i$ is missing. The site-wise RMSE is
\begin{equation}
\operatorname{RMSE}_{\mathrm{site}}(i) =
\left(
\frac{1}{|\mathcal{T}_i|}\sum_{t\in\mathcal{T}_i}(\hat{x}_{t,i}-x_{t,i})^2
\right)^{1/2},
\end{equation}
and the mean site-wise RMSE corresponds to its area-weighted mean over the set $\mathcal{I}$ of cells with $|\mathcal{T}_i|>0$
\begin{equation}
\operatorname{MeanSitewiseRMSE} =
\frac{\sum_{i\in\mathcal{I}} w_i\,\operatorname{RMSE}_{\mathrm{site}}(i)}{\sum_{i\in\mathcal{I}} w_i}.
\end{equation}
Further information on these error measurements can be found in Appendix~\ref{app:metrics}, and maps of the spatially resolved vector $\operatorname{RMSE}_{\mathrm{site}}(i)$ are presented in Section~\ref{sec:results}. 
 
\subsection{Uncertainty Quantification}
\label{sec:uq}
 
We compute $95\%$ confidence intervals for paired differences in monthly spatial RMSE using a nonparametric block bootstrap. Contiguous 12-month blocks are sampled with replacement to form synthetic time series matching the original length, over which mean paired differences are computed across $10{,}000$ iterations. Block sampling accounts for temporal autocorrelation in reconstruction error caused by seasonal cycles and long-term trends in observational coverage. The same procedure gives a confidence interval for the slope of the paired monthly difference against observed fraction (Section~\ref{sec:coverage}), with the slope recomputed by ordinary least squares in each resample.

\section{Data}
\label{sec:data}

\subsection{CMIP6 CESM2 Ensemble}

We use monthly near-surface air temperature (\textit{tas}) from historical simulations of the sixth phase of the Coupled Model Intercomparison Project \cite{eyring_cmip6_2016}. We restrict the ensemble to the Community Earth System Model version 2 family \cite{Danabasoglu2020}. Fields were accessed from the Google Cloud CMIP6 Zarr archive, filtered to \texttt{activity\_id} = \texttt{CMIP}, \texttt{institution\_id} = \texttt{NCAR}, \texttt{table\_id} = \texttt{Amon}, and \texttt{experiment\_id} = \texttt{historical}. We regrid each field with xESMF \cite{zhuang_xesmf} using bilinear interpolation onto a uniform $5^\circ$ latitude--longitude grid spanning $-90^\circ$ to $90^\circ$ in latitude and $0^\circ$ to $355^\circ$ in longitude. This gives $37 \times 72 = 2664$ grid cells. All fields cover January 1850 to December 2014, which corresponds to $T = 1980$ monthly timesteps.

We use twenty realizations spanning two distinct sources of variability. Realizations \texttt{r1} to \texttt{r11} of the standard CESM2 configuration are initial-condition ensemble members. They share identical model physics and external forcing, differing only in the atmospheric perturbation applied at initialization. Therefore, differences among them arise solely from internal variability. The remaining nine realizations originate from three alternative model configurations (three members each). All four configurations share the same finite-volume dynamical core. CESM2-FV2 runs the CAM6 atmosphere at approximately $2^\circ$ horizontal resolution instead of the standard $1^\circ$. CESM2-WACCM replaces CAM6 with the WACCM6 atmosphere, which features 70 vertical levels extending into the lower thermosphere and interactive middle-atmosphere chemistry. CESM2-WACCM-FV2 combines the WACCM6 atmosphere with the coarser $2^\circ$ resolution. A full list is provided in Appendix~\ref{app:realizations}.

Because all realizations are regridded to a common $5^\circ$ grid, native resolution differences manifest primarily as an effective smoothing of the source fields. Incorporating the alternative configurations exposes the model to both structural and internal variability during training. Both held-out realizations are drawn from the standard configuration, ensuring our evaluation measures generalization to an unseen realization within a known model family rather than generalization across fully independent Earth system models.

\subsection{Train, Validation, and Test Partition}

We partition by realization rather than by timestep to prevent data leakage, ensuring no held-out field shares a simulation trajectory with any training field. Realization \texttt{r10} is held out for validation, and the model checkpoint with the lowest validation error is selected. Realization \texttt{r11} is held out strictly for the final evaluation presented in Section~\ref{sec:results}.

All fields are standardized on a per-calendar-month and per-grid-cell basis. For month $m$ and cell $i$, we compute a mean $\mu_{m,i}$ and standard deviation $\sigma_{m,i}$ exclusively from the eighteen training realizations, transforming each value as $(x_{t,i} - \mu_{m,i}) / \sigma_{m,i}$. This step eliminates the seasonal cycle and equalizes variance across cells, such that a value of zero represents the local climatological mean for that month. These normalization statistics are applied unchanged to \texttt{r10} and \texttt{r11}. Every baseline method evaluated uses identical preprocessing and statistics.

\subsection{HadCRUT5 Observational Mask}
\label{sec:mask}

We derive binary observation masks from the HadCRUT5 non-infilled anomaly product \cite{Morice2021}, which reports values only where observational records exist. A cell is classified as observed at month $t$ if HadCRUT5 reports a value there. We regrid the binary mask to the CMIP6 $5^\circ$ grid via nearest-neighbor selection and restrict it to the 1980 months common to both records. Each realization is masked with this identical time-indexed sequence, so that month $t$ of every realization reproduces the observational coverage available in that calendar month.

\begin{figure}[htbp]
\centering
\includegraphics[width=\linewidth]{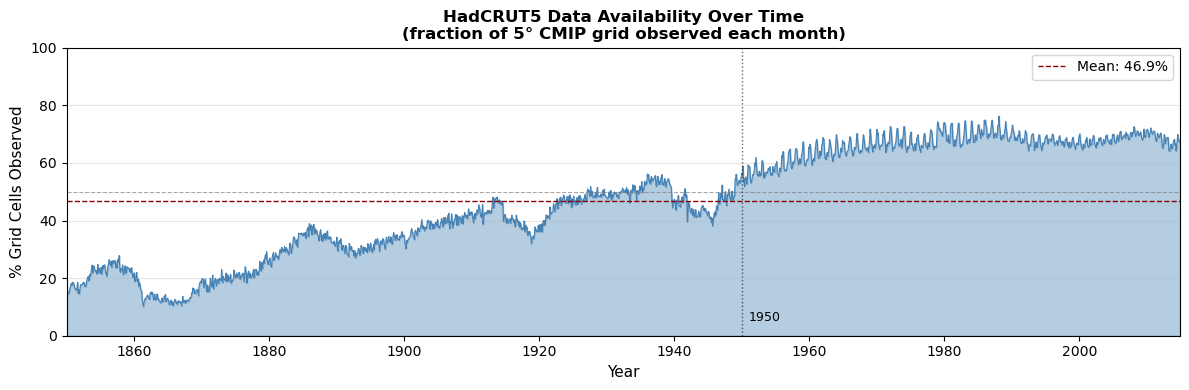}
\caption{Fraction of the $5^\circ$ CMIP6 grid observed in HadCRUT5 for each month from 1850 to 2014. Coverage drops are visible in the 1860s, around 1918, and during the 1940s. After 1950, a distinct annual cycle emerges due to higher polar observation frequency during local summer months.}
\label{fig:coverage}
\end{figure}

\begin{figure}[htbp]
\centering
\includegraphics[width=\linewidth]{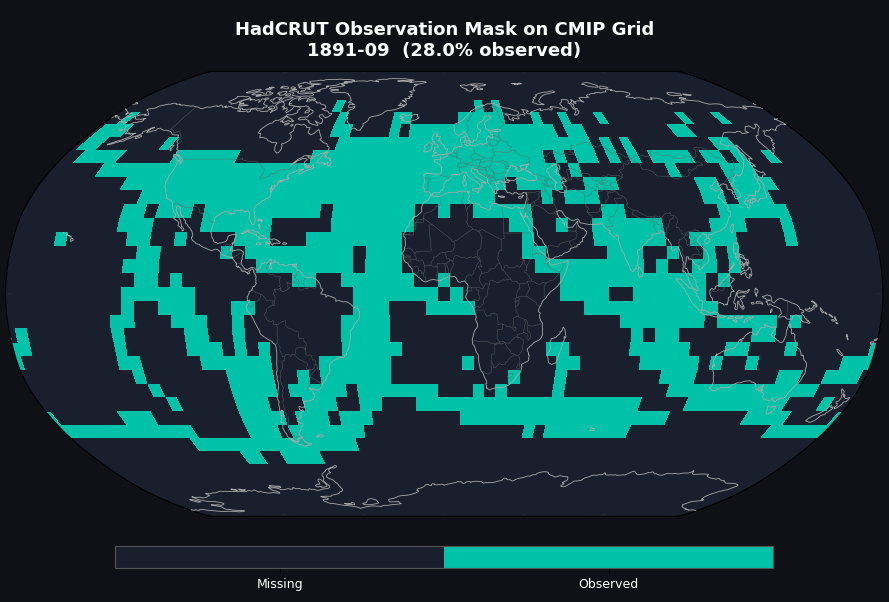}
\caption{HadCRUT5 observation mask for September 1891 after nearest-neighbor regridding to the CMIP6 $5^\circ$ grid. Teal cells denote observed locations; dark cells denote missing values. Only 28.0\% of cells are observed at this timestep, concentrated primarily over Europe, eastern North America, and commercial shipping routes.}
\label{fig:hadcrut_mask}
\end{figure}

Coverage is sparse and strongly non-stationary across the record. Averaged over the entire period, $46.9\%$ of cells are observed. The observed fraction rises from $21.3\%$ in the 1850s to $68.9\%$ in the decade ending 2014, with a monthly minimum of $10.2\%$ and maximum of $76.3\%$ (Figure~\ref{fig:coverage}). Coverage is also spatially non-uniform due to persistent gaps that cover the Southern Ocean, Antarctica, and the central Arctic throughout the record, while nineteenth-century gaps extend across large portions of Africa, South America, and the Pacific Ocean (Figure~\ref{fig:hadcrut_mask}).

Two subsets of cells require special consideration during evaluation. Of the 2664 grid cells, $330$ are never observed in any month. These lie predominantly in Antarctica and the high Arctic. GraphEM, which estimates each location's covariance from that location's own record, cannot produce estimates at these cells, and the remaining methods must infer them entirely from observations elsewhere. Conversely, $36$ cells are observed in every month. They are concentrated over western and central Europe, the eastern United States, European Russia, western Siberia, and Iceland, representing the regions with the longest continuous instrumental records. Because spatial error metrics are evaluated strictly over unobserved cells, these $36$ cells never enter the evaluation metric.

\subsection{Auxiliary Fields}

Surface elevation is retrieved from the Open-Meteo Elevation API \cite{openmeteo} at the center coordinate of each grid cell. We derive a binary land indicator from elevation, yielding a global land area fraction of $32.1\%$. Both fields are static over time. Elevation serves as a node feature, whereas the land indicator is incorporated both as a node feature and to construct land--ocean boundary transition features along graph edges, as detailed in Section~\ref{sec:methods}.

\section{Results}
\label{sec:results}

All results are reported on the held-out realization \texttt{r11}. Every method is evaluated against the same ground truth, with the same time-indexed masks, the same seasonal standardization, and the same metrics. Neither learned model was tuned on \texttt{r11}. Checkpoints for both deep learning implementations were selected on \texttt{r10}.

The two special subsets of cells identified in Section~\ref{sec:mask} are handled as follows. GraphEM produces no estimate at the 330 never-observed cells, so the four-way comparison uses the remaining 2334. Results at the never-observed cells, for the three methods that do produce estimates there, are summarized briefly at the end of Section~\ref{sec:spatial}. The 36 always-observed cells have no defined site-wise RMSE for any method, so comparisons across locations use 2298 cells.

\subsection{Comparison of Methods}

On the matched set of 2334 cells, our model attains a spatial RMSE of 0.997\,K and a mean site-wise RMSE of 0.648\,K (Table~\ref{tab:hadcrut_matched}). This represents a 10.1\% reduction in spatial RMSE compared to LaMa, 31.2\% compared to kriging, and 32.2\% compared to GraphEM. In mean site-wise RMSE, the reductions are 2.4\%, 26.4\%, and 11.3\%, respectively. The block bootstrap described in Section~\ref{sec:uq} yields a mean paired difference against LaMa of 0.111\,K with a 95\% confidence interval of $[0.103, 0.120]$\,K. Our model achieved lower error in all $10{,}000$ resamples.

\begin{table}[htbp]
\centering
\caption{Reconstruction error on CESM2 \texttt{r11} under HadCRUT5 masks,
matched evaluation set of 2334 cells over 1980 months. RMSE in kelvin.}
\label{tab:hadcrut_matched}
\begin{tabular}{lrr}
\toprule
Method & SpatialRMSE & MeanSitewiseRMSE \\
\midrule
GATv2 & \textbf{0.997} & \textbf{0.648} \\
LaMa         & 1.108          & 0.664 \\
GraphEM      & 1.471          & 0.731 \\
Kriging      & 1.450          & 0.880 \\
\bottomrule
\end{tabular}
\end{table}

The two statistical baselines rank differently under the two metrics. GraphEM and kriging perform similarly in spatial RMSE (1.471\,K vs.\ 1.450\,K) but differ substantially in mean site-wise RMSE (0.731\,K vs.\ 0.880\,K). Spatial RMSE weights every month equally and is therefore dominated by the sparsely observed early record, whereas site-wise RMSE pools errors over time at each location and rewards local temporal consistency. GraphEM demonstrates higher location-specific consistency, while both reconstruct a typical month with similar aggregate accuracy.

\subsection{Spatial Distribution of Error}
\label{sec:spatial}

\begin{figure}[htbp]
\centering
\includegraphics[width=\linewidth]{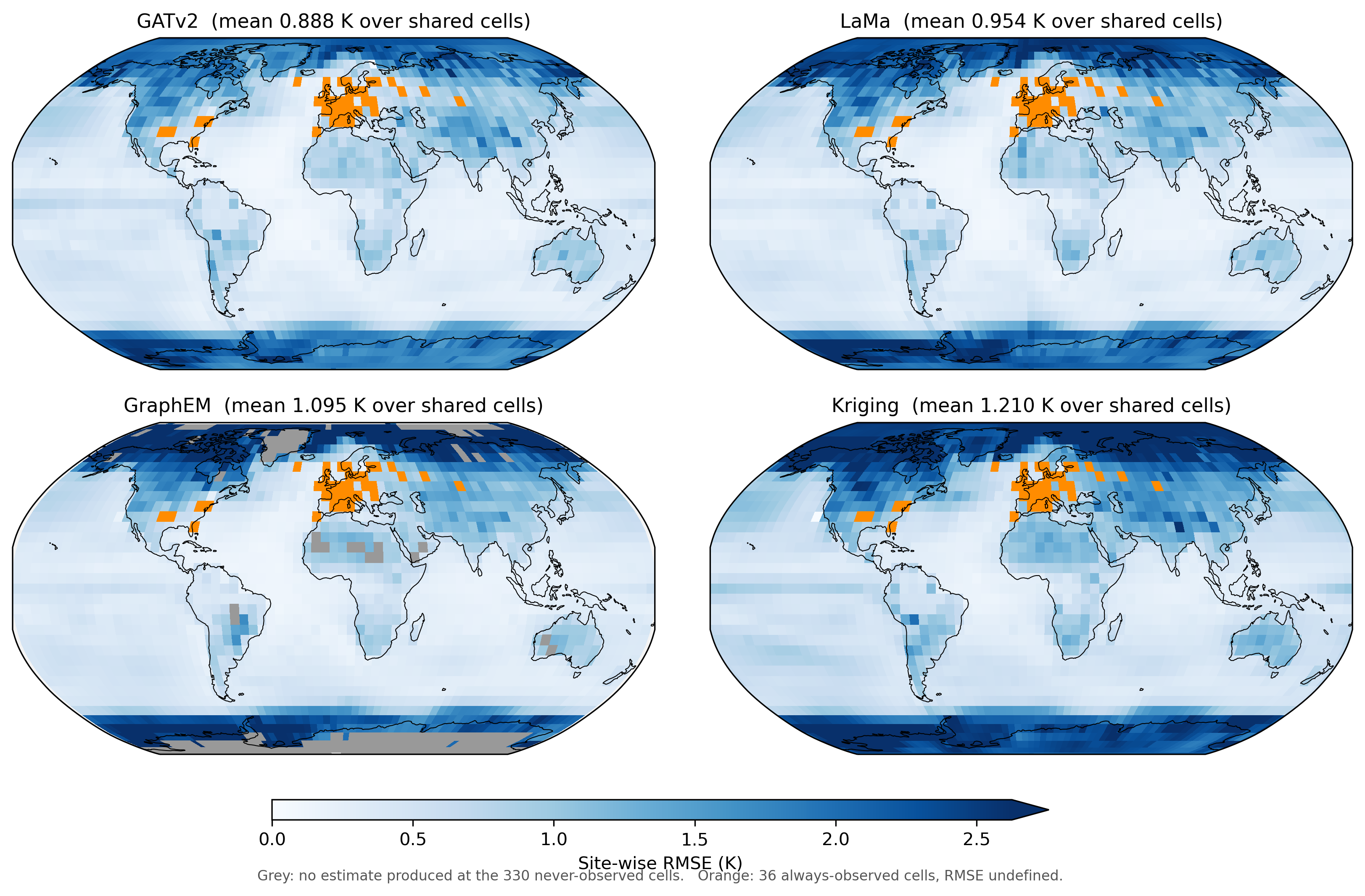}
\caption{Site-wise RMSE on CESM2 \texttt{r11}, displayed on a common color scale saturating at the 92nd percentile to preserve mid-latitude visual structure. Panel means are calculated over cells for which every method returns a value. Grey cells denote never-observed locations where GraphEM produces no estimate; GATv2, LaMa, and kriging predictions are shown for these cells. Orange cells denote always-observed locations where site-wise RMSE is undefined.}
\label{fig:sitewise_maps}
\end{figure}

For all four methods, site-wise RMSE is largest in the polar regions and smallest in the tropics and subtropical oceans (Figure~\ref{fig:sitewise_maps}), consistent with prior evaluations of kriging, PConv, and LaMa \cite{Kadow2020, Bochow2025}. Kriging errors are also elevated over the Northern Hemisphere continental interiors, including Siberia, central Asia, and North America. The performance advantage of our model over LaMa varies by latitude (Table~\ref{tab:regional}): it reaches 13.8\% in the Arctic and 9.6\% in the Antarctic, compared to 3.5\% in the southern mid-latitudes, while in the tropics LaMa's error is 8.5\% lower. The margin is largest where observational coverage is sparsest and where the latitude--longitude grid deviates most from great-circle geometry.

Our model achieves the lowest error in four of the five latitude bands. The exception is the tropics, where LaMa reaches 0.389\,K and GraphEM 0.407\,K compared to 0.422\,K for GATv2. The tropics combine a spatially smooth temperature field with dense observational coverage which is the operational regime where covariance-based and image-based methods perform best. In the Arctic, where coverage is minimal, GraphEM reaches 2.698\,K compared to 1.822\,K for GATv2.

\begin{table}[htbp]
\centering
\caption{Mean site-wise RMSE (K) by latitude band (unweighted within each band) over the 2298 cells evaluated by all methods.}
\label{tab:regional}
\begin{tabular}{lrrrr}
\toprule
Region & GATv2 & LaMa & GraphEM & Kriging \\
\midrule
Arctic ($60^\circ$--$90^\circ$N)         & \textbf{1.822} & 2.113 & 2.698 & 2.620 \\
N mid-latitude ($30^\circ$--$60^\circ$N) & \textbf{0.810} & 0.831 & 0.845 & 1.146 \\
Tropics ($30^\circ$S--$30^\circ$N)       & 0.422 & \textbf{0.389} & 0.407 & 0.544 \\
S mid-latitude ($30^\circ$--$60^\circ$S) & \textbf{0.437} & 0.453 & 0.526 & 0.612 \\
Antarctic ($60^\circ$--$90^\circ$S)      & \textbf{2.068} & 2.288 & 2.439 & 2.547 \\
\bottomrule
\end{tabular}
\end{table}

\begin{figure}[htbp]
\centering
\includegraphics[width=\linewidth]{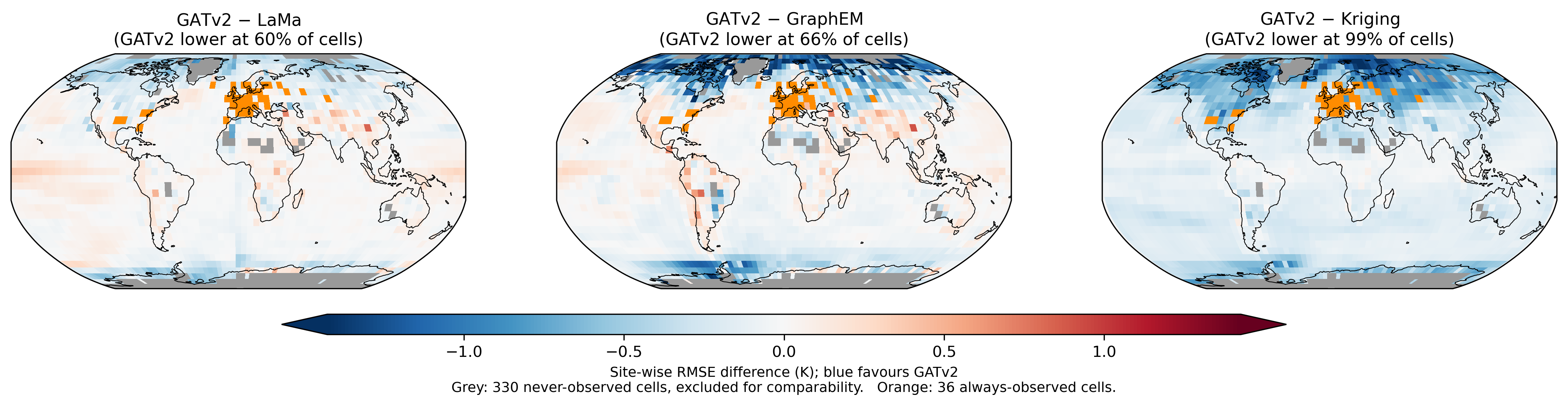}
\caption{Difference in site-wise RMSE between GATv2 and each baseline ($\Delta_i = \mathrm{RMSE}^{\mathrm{GATv2}}_{\mathrm{site}}(i) - \mathrm{RMSE}^{B}_{\mathrm{site}}(i)$). Blue shading indicates lower GATv2 error. The color scale is symmetric about zero and saturates at the 99th percentile of $|\Delta_i|$. Grey cells denote never-observed locations, excluded from all three comparisons so that each panel covers the same cells; orange cells denote always-observed locations where site-wise RMSE is undefined.}
\label{fig:diffmaps}
\end{figure}

Figure~\ref{fig:diffmaps} displays site-wise RMSE differences relative to each baseline. Our model achieves lower error across 60.2\% of evaluated cells compared to LaMa, 66.1\% relative to GraphEM, and 98.5\% relative to kriging. The improvement over kriging is near-universal, whereas the aggregate advantage over GraphEM stems from substantial gains in sparse regions rather than minor global improvements.

Relative to LaMa, error reductions are concentrated at high northern latitudes (northern Siberia, the Canadian Arctic, and Alaska) and across the Southern Ocean. LaMa performs better along zonal bands across the tropical oceans and at scattered, well-observed land cells in Europe, the Middle East, and central Asia. Across the tropics and mid-latitudes, differences of either sign are mostly small. Relative to GraphEM, reductions exceed 1\,K across the Arctic and the Southern Ocean, though GraphEM performs better across much of the tropical oceans and at parts of the South American and Asian landmasses. Relative to kriging, our model is more accurate almost everywhere, with the largest reductions at high northern latitudes. At the 330 never-observed cells, our model attains a spatial RMSE of 1.522\,K, compared to 1.754\,K for LaMa and 2.094\,K for kriging (reductions of 13.2\% and 27.3\%).

\subsection{Dependence on Observational Coverage}
\label{sec:coverage}

\begin{figure}[htbp]
\centering
\includegraphics[width=\linewidth]{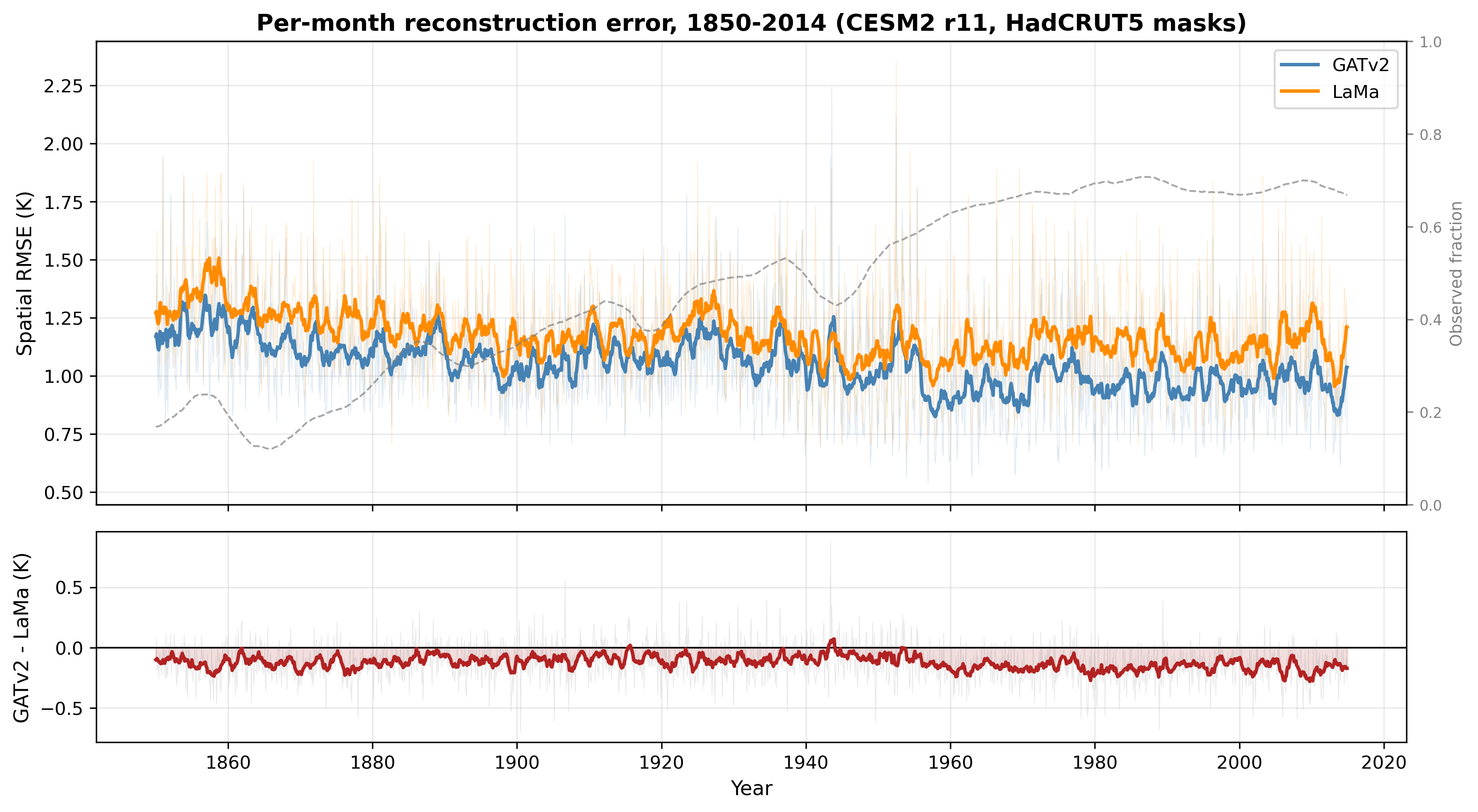}
\caption{Per-month spatial RMSE on CESM2 \texttt{r11}. \textit{Top:} Monthly values in pale shading with 12-month rolling means overlaid. The dashed line denotes the five-year smoothed fraction of observed grid cells (right axis). \textit{Bottom:} 12-month smoothed error difference; negative values indicate lower GATv2 error.}
\label{fig:temporal}
\end{figure}

The two deep learning models respond differently to temporal coverage increases (Figure~\ref{fig:temporal}). Our model's spatial RMSE improves from 1.219\,K in the first decade to 0.978\,K in the final decade (a 19.8\% reduction), whereas LaMa's error improves less (1.350\,K to 1.155\,K, a 14.4\% reduction). Consequently, the performance gap between the two widens from 0.130\,K to 0.177\,K over the record. Our model's error is lower than LaMa's throughout nearly the entire record (Figure~\ref{fig:temporal}, bottom), and the gap is largest after the mid-1950s, when observed coverage rises above 60\%. Averaged over the full record, our model's monthly spatial RMSE is $0.111$\,K lower than LaMa's. A 12-month block bootstrap (Section~\ref{sec:uq}) gives a 95\% confidence interval of $[0.103,\allowbreak 0.120]$\,K for this difference, and our model is lower in all $10{,}000$ resamples.

\begin{figure}[htbp]
\centering
\includegraphics[width=\linewidth]{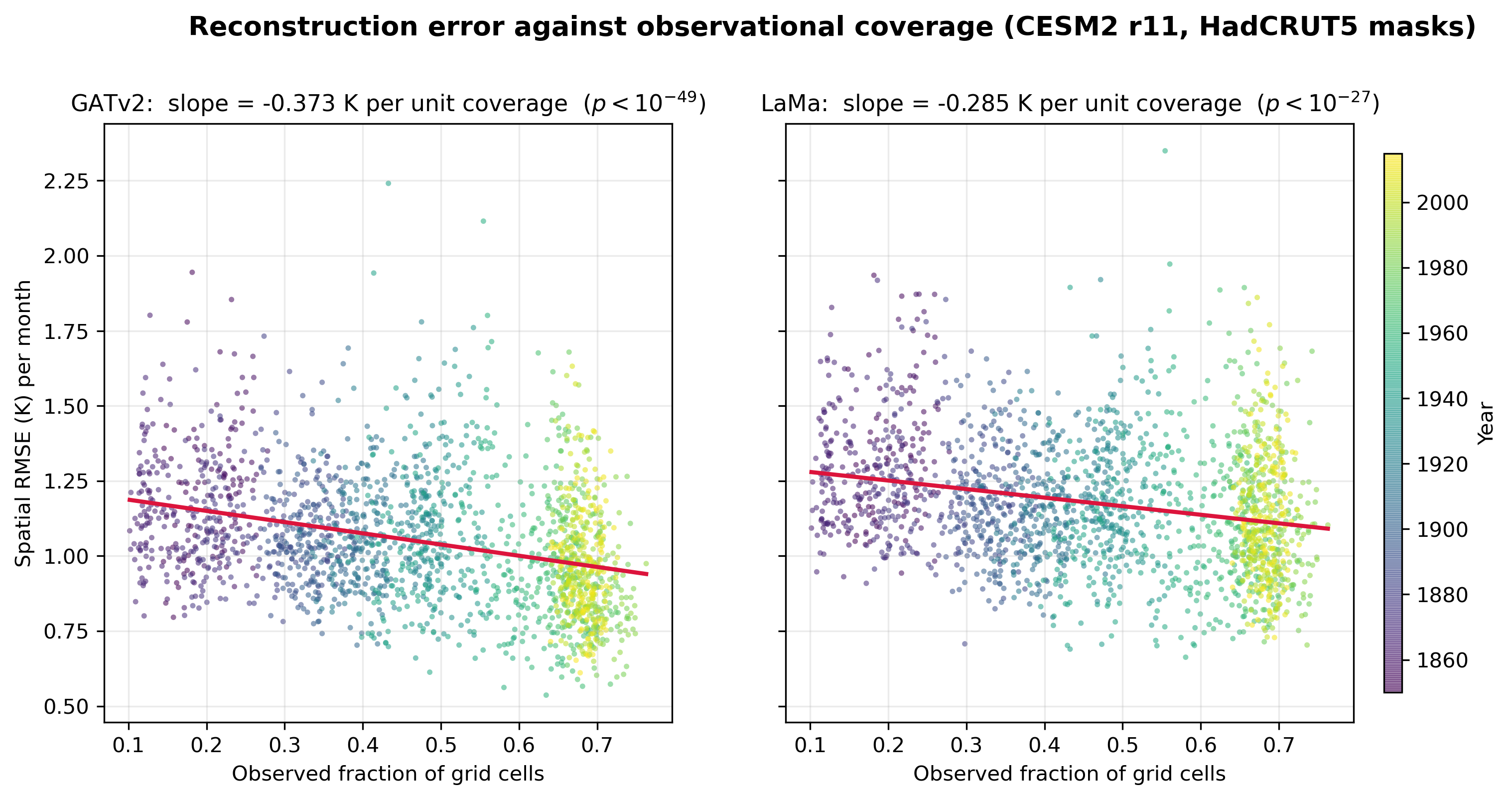}
\caption{Per-month spatial RMSE versus observed cell fraction for GATv2 (left) and LaMa (right). Each point denotes one month, colored by year. Solid lines represent ordinary least squares fits.}
\label{fig:coverage_scatter}
\end{figure}

Linear regression of monthly spatial RMSE against observed cell fraction confirms that the gap widens with coverage (Figure~\ref{fig:coverage_scatter}). Our model exhibits a slope of $-0.373$\,K per unit coverage ($p < 10^{-49}$), whereas LaMa exhibits a shallower slope of $-0.285$\,K per unit coverage ($p < 10^{-27}$). Both models' errors decrease as observational density increases, with our model's decreasing faster. Regressing the monthly difference between the two models' errors (LaMa minus our model) on observed fraction gives a slope of $+0.088$\,K per unit coverage, with a 12-month block-bootstrap 95\% confidence interval of $[0.047,\allowbreak 0.129]$\,K (Section~\ref{sec:uq}). The gap between the two models therefore widens significantly as coverage increases.

\subsection{Ablations}

Table~\ref{tab:ablation} isolates the contribution of individual graph components. The complete configuration reduces spatial RMSE by $5.7\%$ relative to a base graph containing only masked fields and great-circle distance. The observation mask channel reduces spatial RMSE by $1.4\%$, elevation adds a $1.5\%$ reduction, land features a further $0.7\%$, and a second attention head provides a $2.2\%$ reduction. Every addition lowers the error, but land features contribute the least.

These feature effects should be contextualized by the sensitivity to graph connectivity reported in Appendix~\ref{app:ksweep}, where increasing neighborhood size $k$ from 4 to 16 yields a $19\%$ spatial RMSE reduction, more than three times the combined effect of all features and the second attention head. The core advantage of this approach stems from physical graph geometry and learned attention rather than auxiliary node/edge features.

\begin{table}[htbp]
\centering
\caption{Ablation of node/edge features and attention heads ($k = 16$, full evaluation set of 2664 cells). Each row accumulates features from the row above. RMSE in kelvin.}
\label{tab:ablation}
\begin{tabular}{lccccrr}
\toprule
Variant & Mask & Elev. & Land & Heads & SpatialRMSE & MeanSitewiseRMSE \\
\midrule
\texttt{baseline}            &            &            &            & 1 & 1.1125 & 0.7173 \\
\texttt{mask}                & \checkmark &            &            & 1 & 1.0968 & 0.7089 \\
\texttt{mask\_elev}          & \checkmark & \checkmark &            & 1 & 1.0806 & 0.7011 \\
\texttt{mask\_elev\_land}    & \checkmark & \checkmark & \checkmark & 1 & 1.0733 & 0.6986 \\
\texttt{mask\_elev\_land\_2h}& \checkmark & \checkmark & \checkmark & 2 & \textbf{1.0496} & \textbf{0.6748} \\
\bottomrule
\end{tabular}
\end{table}

\subsection{Computational Cost}
\label{sec:compute}

Our model utilizes 57{,}729 trainable parameters compared to 27{,}090{,}899 in the LaMa generator ($469\times$ fewer) and requires $6.4\times$ fewer optimizer steps. Once trained, single-realization inference requires 4.2 seconds for GATv2, compared to 15.7 minutes for kriging and 32.6 minutes for GraphEM. Reconstructing ten additional realizations requires approximately 42 seconds in total for GATv2, versus 2.6 hours for kriging and 6.6 hours for GraphEM (platform runtime specifications and hardware details are documented in Appendix~\ref{app:runtime}).

\section{Conclusion}
\label{sec:conclusion}

We have presented a distance-aware graph attention network for reconstructing missing values in gridded temperature fields, evaluating its performance against ordinary kriging, GraphEM, and a deep inpainting model under observationally realistic missingness. Representing the grid as a $k$-nearest-neighbor graph under great-circle distance rather than as a regular pixel lattice establishes a physical notion of locality that respects spherical geometry, naturally wraps the dateline, and seamlessly accommodates edge as well as node covariates.

On a held-out CESM2 realization masked with the time-indexed HadCRUT5 record, our method attains lower spatial RMSE than all three baselines, reducing error by 10.1\% relative to LaMa and by approximately 31\% and 32\% relative to kriging and GraphEM. A nonparametric block bootstrap over monthly errors places the performance advantage over LaMa at $0.111$\,K ($95\%$ CI: $[0.103, 0.120]$\,K). This margin is largest in the Arctic and Antarctic, where observational coverage is sparsest and latitude--longitude grids deviate most from great-circle geometry. It reverses in the tropics, where LaMa and GraphEM achieve higher accuracy. The model also produces the most accurate estimates at the 330 grid cells never observed in HadCRUT5.

Two core insights emerge regarding how these models operate. First, the two deep learning methods respond differently to observational density. Both improve as the observing network expands over time, but our model's error decreases significantly faster, widening the performance gap between the two by more than a third between the 1850s and 2000s. LaMa therefore makes less use of each new observation as compared to our model. Second, feature ablations demonstrate that auxiliary physical covariates and a second attention head together provide only modest gains ($5.7\%$ spatial RMSE reduction), with land--ocean features contributing the least ($0.7\%$). In contrast, varying the neighborhood size $k$ moves the error more than three times as much. The model's primary advantage therefore stems from the geometric construction of the graph and learned attention over spatial neighborhoods rather than auxiliary physical features.

Several limitations warrant consideration. First, evaluating on a single held-out realization from one model family measures generalization to unseen realizations within a known physics package rather than across independent Earth system models. Second, skill is evaluated against climate simulation ground truth rather than real-world observational data, leaving open how the network responds to real instrumental artifacts. Third, fields were regridded bilinearly rather than conservatively, sampling source resolution rather than preserving total heat content. While all baseline comparisons remain strictly controlled under identical inputs, this choice may influence baseline error levels.

Because the model's advantage stems from the geometric construction of the graph rather than from the physical covariates, and because its training objective imposes no Gaussian assumption, this formulation extends naturally to other gridded climate records. Applying it to variables with localized, non-Gaussian spatial statistics, such as precipitation, wind speed, and cloud cover, will determine whether spherical graph representations generalize beyond smooth, strongly autocorrelated temperature fields, and, given the sensitivity of skill to neighborhood size, how the optimal neighborhood scales with each variable's spatial correlation length. Thus, a further step is to extend the approach to the $1^\circ$ to $0.25^\circ$ resolutions of modern Earth system models and reanalyses. Since the spherical graph approximates a continuous surface, theory for graph neural networks suggests that a model trained at one resolution should transfer to another~\cite{ruiz,wang}, which would substantially reduce the cost of high-resolution reconstruction and remove the need to regrid to a coarse common grid.

\section*{Acknowledgments}

This work used Bridges-2 at the Pittsburgh Supercomputing Center through allocation cis261081p from the Advanced Cyberinfrastructure Coordination Ecosystem: Services \& Support (ACCESS) program, which is supported by U.S. National Science Foundation grants~\#2138259, \#2138286, \#2138307, \#2137603, and~\#2138296.

\section*{Code and Data Availability}
\label{sec:code}

Code for the model, baselines, and evaluation pipeline will be made publicly available on GitHub upon publication. CMIP6 climate model fields were obtained from the Google Cloud CMIP6 Zarr archive, and HadCRUT5 missingness masks were sourced from the Met Office Hadley Centre. Surface elevation data were retrieved from the Open-Meteo Elevation API. The LaMa baseline implementation derives from the climate adaptation by Bochow et al.~\cite{Bochow2025}; all architectural modifications and hyperparameter settings will be documented in the code repository.

\bibliographystyle{plain}
\bibliography{references}

\appendix

\section{CMIP6 Realizations}
\label{app:realizations}

All fields are \texttt{historical} experiment runs of the \texttt{i1p1f1} variant.

\begin{table}[htbp]
\centering
\caption{CESM2-family realizations used in this study.}
\label{tab:realizations}
\begin{tabular}{llc}
\toprule
Configuration & Realizations & Split \\
\midrule
CESM2           & \texttt{r1}--\texttt{r9} & Train \\
CESM2-FV2       & \texttt{r1}--\texttt{r3} & Train \\
CESM2-WACCM     & \texttt{r1}--\texttt{r3} & Train \\
CESM2-WACCM-FV2 & \texttt{r1}--\texttt{r3} & Train \\
CESM2           & \texttt{r10}             & Validation \\
CESM2           & \texttt{r11}             & Test \\
\bottomrule
\end{tabular}
\end{table}

\section{Properties of the Evaluation Metrics}
\label{app:metrics}

The two metrics aggregate the same squared errors in opposite orders. Spatial RMSE takes the square root within each month and then averages across months. Site-wise RMSE pools across months at each location and takes the square root afterwards. Because the square root is a concave function, pooling before taking the root yields a smaller value by Jensen's inequality, which is why spatial RMSE exceeds mean site-wise RMSE by a factor of roughly 1.5 to 2 for every method. They should be read as complementary summaries rather than as independent estimates of a single quantity. Spatial RMSE measures how accurately an individual month is recovered, whereas mean site-wise RMSE measures how reliable a given location is across the temporal record.

Two properties of the site-wise metric are worth noting. Every location enters the mean site-wise RMSE with equal weight, up to the area factor $w_i$, irrespective of $|\mathcal{T}_i|$. A cell missing in only a handful of months therefore contributes as much as one missing throughout the record. Consequently, estimates at well-observed locations rest on few samples and are correspondingly noisier. Locations that are never missing have an undefined site-wise RMSE and are excluded. Both metrics use the identical mask for every method compared, so these properties affect all evaluated methods equally.

\section{GATv2 Hyperparameters}
\label{app:hyperparameters}
 
Table~\ref{tab:hyperparameters} lists the architecture and training settings of the graph model used for all main text results. The same settings are used for every variant in the ablation study (Table~\ref{tab:ablation}) and the connectivity sweep (Appendix~\ref{app:ksweep}), apart from the input features, the number of attention heads, and $k$, respectively.
 
\begin{table}[htbp]
\centering
\caption{Architecture and training settings of the GATv2 model.}
\label{tab:hyperparameters}
\begin{tabular}{ll}
\toprule
Setting & Value \\
\midrule
\multicolumn{2}{l}{\textit{Architecture}} \\
Node input features              & 4 (masked temperature, mask, elevation, land) \\
Edge input features              & 2 (distance, coastline crossing) \\
Hidden width                     & 64 \\
GATv2 layers ($L$)               & 3 \\
Attention heads                  & 2 (outputs averaged) \\
Edge MLP                         & 2 linear layers, width 16, ReLU \\
Activation                       & ELU \\
Normalization                    & Layer normalization with residual connection \\
Dropout                          & 0.3 (attention coefficients and hidden features) \\
Neighbors per node ($k$)         & 16 \\
Trainable parameters             & 57{,}729 \\
\midrule
\multicolumn{2}{l}{\textit{Training}} \\
Optimizer                        & Adam, initial learning rate $5 \times 10^{-4}$ \\
Learning-rate schedule           & Halved after 5 validation checks without improvement \\
Batch size                       & 32 monthly fields \\
Gradient accumulation            & 2 batches (effective batch size 64) \\
Epochs                           & 100 \\
Optimizer steps                  & 55{,}700 (557 per epoch) \\
Training samples                 & 35{,}640 monthly fields (18 realizations $\times$ 1980 months) \\
Loss                             & Mean squared error at unobserved cells \\
Validation                       & Every 5 epochs on \texttt{r10} \\
Checkpoint selection             & Lowest validation RMSE \\
Mixed precision                  & Yes (automatic loss scaling) \\
\bottomrule
\end{tabular}
\end{table}

\section{Graph Connectivity Sweep}
\label{app:ksweep}

We vary the number of nearest neighbors $k$, holding all other architectural and optimization settings fixed. Each variant is trained under the same set-up detailed in Section~\ref{sec:methods} and selected on the held-out validation realization \texttt{r10}. Edge counts are those of the symmetrized graph described in Section~\ref{sec:graph}.

\begin{table}[htbp]
\centering
\caption{Effect of graph connectivity $k$ on reconstruction error. Test metrics are evaluated on \texttt{r11} for the checkpoint selected on \texttt{r10}.}
\label{tab:ksweep}
\begin{tabular}{lrrrrr}
\toprule
$k$ & Edges & Val RMSE & SpatialRMSE & MeanSitewiseRMSE & Train (h) \\
\midrule
4   & 11{,}376 & 0.7907 & 1.2995 & 0.7547 & 2.0 \\
8   & 23{,}760 & 0.7396 & 1.1202 & 0.7030 & 3.2 \\
16  & 46{,}800 & 0.6879 & 1.0496 & 0.6748 & 5.6 \\
32  & 95{,}040 & \textbf{0.6701} & \textbf{1.0252} & \textbf{0.6620} & 10.5 \\
\bottomrule
\end{tabular}
\end{table}

Validation and test metrics order the variants identically which demonstrates that hyperparameter selection on validation data generalizes directly to the test fields. Error decreases monotonically with $k$ but exhibits diminishing returns. Successive doublings of $k$ reduce spatial RMSE by 0.179\,K, 0.071\,K, and 0.024\,K, while training time increases by a factor of 1.6 to 1.9 at each step.

We select $k = 16$ for all main text evaluations. Increasing connectivity to $k = 32$ reduces spatial RMSE by an additional $0.024$\,K ($2.3\%$), but increases training time from 5.6 to 10.5 hours (an $88\%$ increase). The $k = 16$ graph achieves performance within $2.3\%$ of the denser graph at nearly half the computational cost. For comparison, Guillot et al.\ \cite{Guillot2015} select an $800$\,km neighborhood radius by cross-validation for GraphEM on a comparable grid, corresponding to an average of $8.42$ neighbors per node.

\section{Feature Ablation Validation Metrics}
\label{app:ablation}
 
Validation RMSE on \texttt{r10} for each ablation variant is listed in Table~\ref{tab:ablation_val}. Validation error decreases monotonically across the five variants, matching the ordering of test spatial RMSE in Table~\ref{tab:ablation}. Adding land features gives the smallest improvement on both splits, reducing validation RMSE by $0.004$\,K and test spatial RMSE by $0.007$\,K. Because both splits improve, the land features appear to provide a small but consistent benefit; however, an effect of this size may be comparable to run-to-run variation from random weight initialization, which we did not quantify, and should be interpreted with caution.
 
\begin{table}[htbp]
\centering
\caption{Validation RMSE on \texttt{r10} for each feature ablation variant.}
\label{tab:ablation_val}
\begin{tabular}{lr}
\toprule
Variant & Val RMSE (K) \\
\midrule
\texttt{baseline}            & 0.7233 \\
\texttt{mask}                & 0.7167 \\
\texttt{mask\_elev}          & 0.7041 \\
\texttt{mask\_elev\_land}    & 0.7001 \\
\texttt{mask\_elev\_land\_2h}& \textbf{0.6879} \\
\bottomrule
\end{tabular}
\end{table}

\section{Baseline Configurations}
\label{app:baselines}

\paragraph{Kriging.} Because the PyKrige implementation \cite{murphy2022pykrige} assumes planar coordinates, spatial inputs are reprojected to the World Cylindrical Equal Area projection (ESRI:54034) using pyproj \cite{snow2022pyproj} prior to kriging. Each month is kriged from all observations available in that month.

\paragraph{GraphEM.} The neighborhood graph is constructed identically to our model (Section~\ref{sec:graph}): $k$-nearest neighbors ($k = 16$) under haversine distance ($R = 6371$\,km), with coincident pole points skipped when choosing neighbors, symmetrized, and with self-loops added. Including self-loops, this yields a mean node degree of $18.6$ and a graph density of $0.0070$ (degree range: $17$--$37$). Both methods operate over identical graph topologies. Imputation uses the expectation–maximization procedure of Guillot et al.\ \cite{Guillot2015}, with the covariance estimated as a Gaussian graphical model whose sparsity pattern is fixed by this graph. The missing values are imputed by their conditional expectation given the observed values, under the package’s default convergence settings.

\paragraph{LaMa.} LaMa was trained for 100 epochs (batch size 10, 3564 batches/epoch). Model checkpoints were selected using the framework's native validation metric (a composite of SSIM and FID) evaluated on \texttt{r10}.

\section{Quantization Floor of the LaMa Input Representation}
\label{app:quantization}

Because LaMa operates on 8-bit greyscale image arrays, standardized temperature fields are quantized into $[0, 255]$ over the range $[-10.106, 14.089]$ standard deviations. One discrete quantization step corresponds to $24.195 / 255 = 0.0949$ standard deviations.

Because LaMa composites predictions as $\hat{x} = m \odot g + (1 - m) \odot x$, observed cells pass through directly, making their reconstruction error strictly a function of quantization precision. Evaluated across all observed test cells, this quantization noise floor equals $0.075$\,K. While small compared to total reconstruction errors ($>1$\,K), it represents a non-zero representation constraint inherent to 8-bit image formats.

\section{Runtime and Compute Specifications}
\label{app:runtime}

GATv2 and LaMa were trained and evaluated on an NVIDIA V100 GPU. Kriging was run on a CPU host and GraphEM on a Google Colab CPU instance. Neither GraphEM nor kriging offers GPU acceleration in their standard implementations.

Inference runtime for learned models reflects forward-pass execution. Data loading, seasonal normalization, and graph construction add approximately 80 seconds for GATv2, dominated by disk I/O operations. Training times are wall-clock totals for the full job, including data loading, validation, and checkpointing.

\begin{table}[htbp]
\centering
\caption{Wall-clock runtime for one realization (1980 monthly fields across $N = 2664$ grid cells).}
\label{tab:runtimes}
\begin{tabular}{llrr}
\toprule
Method & Platform & Training & Reconstruction \\
\midrule
GATv2 & V100 GPU   & 5.6 h         & 4.2 s \\
LaMa         & V100 GPU   & 25.8 h        & $\approx$2.3 min \\
GraphEM      & CPU        & ---           & 32.6 min \\
Kriging      & CPU        & ---           & 15.7 min \\
\bottomrule
\end{tabular}
\end{table}

\end{document}